*Article*

# Learning a Dilated Residual Network for SAR Image Despeckling

**Qiang Zhang [1], Qiangqiang Yuan [1,\*], Jie Li [1,\*], Zhen Yang [2], Xiaoshuang Ma [3]**

1. School of Geodesy and Geomatics, Wuhan University, Wuhan 430079, China;
   zhang_qiang@whu.edu.cn(Q.Z.); qqyuan@sgg.whu.edu.cn(Q.Y.); aaronleecool@whu.edu.cn(J.L.)
2. School of Resource and Environmental Science, Wuhan University, Wuhan 430079, China;
   SendimageYZ@whu.edu.cn (Z.Y.)
3. School of Resources and Environmental Engineering, Anhui University, Hefei 230000, China;
   mxs.88@whu.edu.cn (X.M.)

\* Correspondence: qqyuan@sgg.whu.edu.cn(Q.Y.), aaronleecool@whu.edu.cn(J.L.); Tel.: +86-159-7217-1792 (Q.Y.)

**Abstract:** In this paper, to break the limit of the traditional linear models for synthetic aperture radar (SAR) image despeckling, we propose a novel deep learning approach by learning a non-linear end-to-end mapping between the noisy and clean SAR images with a dilated residual network (SAR-DRN). SAR-DRN is based on dilated convolutions, which can both enlarge the receptive field and maintain the filter size and layer depth with a lightweight structure. In addition, skip connections and residual learning strategy are added to the despeckling model to maintain the image details and reduce the vanishing gradient problem. Compared with the traditional despeckling methods, the proposed method shows superior performance over the state-of-the-art methods on both quantitative and visual assessments, especially for strong speckle noise.

## 1. Introduction

Synthetic aperture radar (SAR) is a coherent imaging sensor, which can access a wide range of high-quality massive surface data. Moreover, with the ability to operate at night and in adverse weather conditions such as thin clouds and haze, SAR has gradually become a significant source of remote sensing data in the fields of geographic mapping, resource surveying, and military reconnaissance. However, SAR images are inherently affected by multiplicative noise, i.e., speckle noise, which is caused by the coherent nature of the scattering phenomena [1]. The presence of speckle severely affects the quality of SAR images, and greatly reduces the utilization efficiency in SAR image interpretation, retrieval, and other applications [2-4]. Consequently, SAR image speckle reduction is an essential preprocessing step and has become a hot research topic.

For the purpose of removing the speckle noise of SAR images, scholars firstly proposed spatial linear filters such as the Lee filter [5], Kuan filter [6] and Frost filter [7]. These methods usually assume that the image filtering result values have a linear relationship with the original image, through searching for a relevant combination of the central pixel intensity in a moving window with a mean intensity of the filter window. Thus, the spatial linear filters achieve a trade-off between balancing in homogeneous areas and a constant all-pass identity filter in edge included areas. The results confirmed that spatial-domain filters which are adept at suppressing speckle noise for some critical features. However, due to the nature of local processing, the spatial linear filter methods often fail to integrally preserve edges and details, which exist the following deficiencies: 1) unable to preserve the average value, especially for the equivalent number of look (ENL) of the original SAR image is small; 2) the powerfully reflective specific targets like points and small surficial features are easily blurred or erased; and 3) speckle noise in dark scene are not removed [8].

Except the spatial-domain filters above, wavelet theory has also been applied into speckle reduction. Starck et al. [9] primarily employed ridgelet transform as a component step, and implemented curvelet sub-bands using a filter bank of the discrete wavelet transform (DWT) filters for image denoising. And for the case of speckle noise, Solbo et al. [10] utilized the DWT of the log-transformed speckled image in homomorphic filtering, which is empirically convergence in a self-adaptive strategy and calculated in the Fourier space. In summary, the major weaknesses of this type of approach are the backscatter mean preservation in homogeneous areas, details preservation, and producing artificial effect into the results such as ring effects [11].

Aimed at overcoming these deficiencies, the nonlocal means (NLM) algorithm [12-14] has provided a breakthrough in detail preservation in SAR image despeckling. The basic idea of the NLM-based methods [12] is that natural images have self-similarity and there are similar patches repeating over and over throughout the whole image. And for SAR image, Deledalle et al. [13] modified the choice of weights, which can be iteratively determined based on both the similarity between noisy patches and the similarity of patches extracted from the previous estimate. Besides, Parrilli et al. [14] used the local linear minimum mean square error (LLMMSE) criterion and undecimated wavelet transform considering the peculiarities of SAR images, allowing for a sparse Wiener filtering representation and an effective separation between original signal and speckle noise through predefined thresholding, which has become one of the most effective SAR despeckling methods. However, the low computational efficiency of the similar patch searching restricts its application.

In addition, the variational-based methods [15-18] have gradually been utilized for SAR image despeckling because of their stability and flexibility, which break through the traditional idea of filters by solving the problem of energy optimization. Then the despeckling task is cast as the inverse problem of recovering the original noise-free image based upon reasonable assumptions or prior knowledge of the noise observation model with log-transform, such as total variation (TV) model [15], sparse representation [16] and so on. Although these variational methods have achieved good reduction of speckle noise, the result is usually dependent on the choice of the model parameters and prior information, and is often time-consuming. In addition, the variational-based methods cannot accurately describe the distribution of speckle noise, which also constraints the performance of speckle noise reduction.

In general, although a lot of SAR despeckling methods have been proposed, they sometimes fail to preserve sharp features in domains of complicated texture, or even create some block artifacts in the speckled image. In this paper, considering that image speckle noise can be expressed more accurately through non-linear models than linear models, and to overcome the above-mentioned limitations of the linear models, we propose a novel deep neural network based approach for SAR image despeckling, learning a non-linear end-to-end mapping between the speckled and clean SAR images by a dilated residual network (SAR-DRN). Our despeckling model employs dilated convolutions, which can both enlarge the receptive field and maintain the filter size and layer depth with a lightweight structure. Furthermore, skip connections are added to the despeckling model to maintain the image details and avoid the vanishing gradient problem. Compared with the traditional despeckling methods in both simulated and real SAR experiments, the proposed approach shows a state-of-the-art performance in both quantitative and visual assessments, especially for strong speckle noise.

The rest of this paper is organized as follows. The SAR image speckling noise degradation model and the related deep convolution neural network method are introduced in Section 2. The network architecture of the proposed SAR-DRN and details of its structure are described in Section 3. Then, the results of the despeckling assessment in both simulated and real SAR image experiments are presented in Section 4. Finally, the conclusions and future research are summarized in Section 5.

## 2. Related Work

*2.1 SAR Image Speckling Noise Degradation Model*

For SAR image, the main reason for the degradation of the image quality is multiplicative speckle noise. Differing from additive white Gaussian noise (AWGN) in nature or hyperspectral image [19]–[20], speckle noise is described by the multiplicative noise model:

$$y = x \cdot n \quad (1)$$

where $y$ is the speckled noise image, $x$ is the clean image, and $n$ represents the speckle noise. It is well-known that, for SAR amplitude image, the speckle follows a *Gamma* distribution [21]:

$$\rho_n(n) = \frac{L^L n^{L-1} \exp(-nL)}{\Gamma(L)} \quad (2)$$

where $L \geq 1$, $n \geq 0$, $\Gamma$ is the *gamma* function, and $L$ is the equivalent number of looks (*ENL*), as defined in (3), which is usually regarded as the quantitative evaluation index for real SAR image despeckling experiments in the homogeneous areas.

$$ENL = \frac{\bar{x}}{\text{var}} \quad (3)$$

where $\bar{x}$ and var respectively represent the image mean and variance.

Therefore, for this non-linear multiplicative noise, choosing a non-linear expression for speckle reduction is an important strategy. In the following, we briefly introduce the use of convolutional neural networks (CNNs) for SAR image despeckling, considering both the low-level features as the bottom level and the output feature representation from the top level of the network.

*2.2 CNNs for SAR Image Despeckling*

With recent advances made by deep learning for computer vision and image processing applications, it has gradually become an efficient tool which has been successfully applied to many computer vision tasks such as image classification, segmentation, object recognition, scene classification and so on [22]–[24]. CNNs can extract the internal and underlying features of images and avoid complex *priori* constraint, organized in the $j$-th feature map $O_j^{(l)}$ ($j = 1, 2, \ldots M^{(l)}$) of $l$-th layer, within which each unit is connected to local patches of the previous layer $O_j^{(l-1)}$ ($j = 1, 2, \ldots M^{(l-1)}$) through a set of weight parameters $W_j^{(l)}$ and bias parameters $b_j^{(l)}$. The output feature map is:

$$L_j^{(l)}(m,n) = F(O_j^{(l)}(m,n)) \quad (4)$$

And

$$O_j^{(l)}(m,n) = \sum_{i=1}^{M^{(l)}} \sum_{u,v=0}^{S-1} W_{ji}^{(l)}(u,v) \cdot L_i^{(l-1)}(m-u, n-v) + b_j^{(l)} \quad (5)$$

where $F(\cdot)$ is the nonlinear activation function, and $O_j^{(l)}(m,n)$ represents as the convolutional weighted sum of the previous layer's results, to the $j$-th output feature map at pixel $(m,n)$. Besides, the special parameters in the convolution layer contain number of output feature maps $j$, and filter kernel size $S \times S$. Particularly, the network parameters $W$ and $b$ need to be regenerated through back-propagation (BP) algorithm and the chain rule of derivation [25].

To ensure that the output of the CNNs is a non-linear combination of the input, due to the relationship between the input data and the output label should usually be a highly nonlinear mapping, a non-linear function is introduced as an excitation function, such as the rectified linear unit (ReLU) is defined as:

$$F(O_j^{(l)}) = \max(0, O_j^{(l)}) \quad (6)$$

After finishing each process of forward propagation, BP algorithm starts to perform for update trainable parameters of networks, to better learn the relationships between label data and reconstructing data. From the top layer of the network to the bottom, BP updates the trainable parameters of $l$-th layer through the outputs of $l+1$-th layer. The partial derivative of loss function with respect to convolution kernels $W_{ji}^{(l)}$ and bias $b_j^{(l)}$ of $l$-th convolution layer is respectively calculated as follows:

$$\frac{\partial L}{\partial W_{ji}^{(l)}} = \sum_{m,n} \delta_j^{(l)}(m,n) \cdot L_j^{(l)}(m-u, y-v) \tag{7}$$

$$\frac{\partial L}{\partial b_j^{(l)}} = \sum_{m,n} \delta_j^{(l)}(m,n) \tag{8}$$

where the error map $\delta_j^{(l)}$ is defined as

$$\delta_j^{(l)} = \sum_j \sum_{u,v=0}^{S-1} W_{ji}^{(l+1)}(u,v) \cdot \delta_j^{(l+1)}(m+u, n+v) \tag{9}$$

And the iterative training rule for updating the networks parameter $W_{ji}^{(l)}$ and $b_j^{(l)}$ is through the gradient descent strategy as follows:

$$W_{ji}^{(l)} = W_{ji}^{(l)} - \alpha \cdot \frac{\partial L}{\partial W_{ji}^{(l)}} \tag{10}$$

$$b_j^{(l)} = b_j^{(l)} - \alpha \cdot \frac{\partial L}{\partial b_j^{(l)}} \tag{11}$$

where $\alpha$ is a preset hyperparameter for the whole network, which is also named learning rate in deep learning framework and controls sampling interval of the trainable parameter.

For natural Gaussian noise reduction, a new method named the feed-forward denoising convolutional neural network (DnCNN) [26] has recently shown excellent performances, in contrast with the traditional methods which employ a deep convolutional neural network. DnCNN employs a 20 convolutional layers structure, a learning strategy of residual learning to remove the latent original image in the hidden layers, and an output data regularization method of batch normalization [27], which can deal with several universal image restoration tasks such as blind or non-blind image Gaussian denoising, single image super-resolution and JPEG image deblocking.

Recently, borrowing the thought of DnCNN model, Chierchia *et al.* [28] also employed a set of convolutional layers named SAR-CNN, along with batch normalization (BN) and ReLU activation function, and a component-wise division residual layer to estimate the speckled image. As an alternative way of dealing with the multiplicative noise of SAR images, SAR-CNN uses the homomorphic approach with coupled logarithm and exponent transforms in combination with a similarity measure for speckle noise distribution. In addition, Wang *et al.* [29] also used a similar structure like DnCNN, with an eight-layers of *Conv-BN-ReLU* block, and replaced residual mean square error (MSE) with a combination of Euclidean loss and total variation loss, which is incorporated into the total loss function to facilitate more smooth results.

**3. Proposed Method**

In this paper, rather than using log-transform [28] or modifying training loss function like [29], we propose a novel network for SAR image despeckling with dilated residual network (SAR-DRN), which is trained in an end-to-end fashion using a combination of dilated convolutions and skip connections with residual learning structure. Instead of relying on pre-determined image a *priori* knowledge or a noise description model, the main superiority of using the deep neural network strategy for SAR image despeckling is that the model can directly acquire and update the network parameters from the training data and the corresponding labels, which needn't manually adjust critical parameters and can automatically learning the complex internal non-linear relations with trainable network parameters from the massive training simulative data.

The proposed holistic neural network model (SAR-DRN) for SAR image despeckling contains seven dilated convolution layers and two skip connections, as illustrated in Figure 1. In addition, the proposed model uses a residual learning strategy to predict the speckled image, which adequately utilizes the non-linear expression ability of deep learning. The details of the algorithm are described in the following.

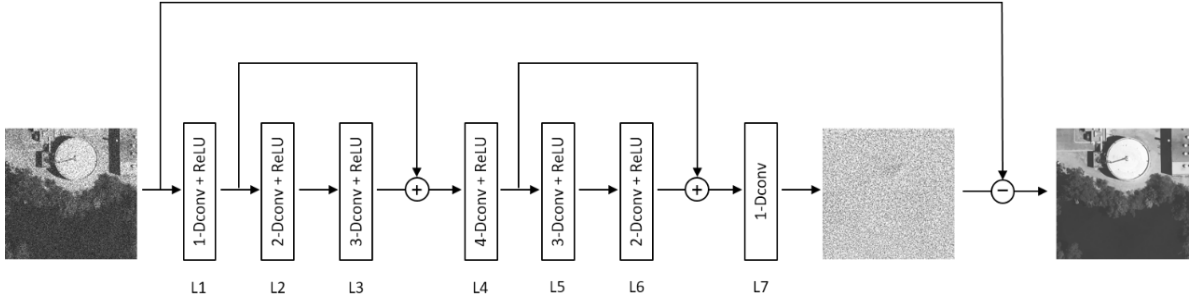

**Figure 1. The architecture of the proposed SAR-DRN.**

*3.1 Dilated Convolutions*

In image restoration problems such as single-image super-resolution (SISR) [30], denoising [31] and deblurring [32], contextual information can effectively facilitate the recovery of degraded regions. In deep convolutional networks, it mainly augments the contextual information through enlarging the receptive field. Generically, there are two ways to achieve this purpose: 1) increasing the network depth; and 2) enlarging the filter size. Nevertheless, as the network depth increases, the accuracy becomes "saturated" and then degrades rapidly. Enlarging the filter size can also lead to more convolution parameters, which greatly increases the calculative burden and training times.

To solve this problem effectively, dilated convolutions were first proposed in [33], which can both enlarge the receptive field and maintain the filter size. Let $C$ be an input discrete 2-dimensional matrix such as image, and let $k$ be a discrete convolution filter of size $(2r+1) \times (2r+1)$. Then the original discrete convolution operator $*$ can be given as

$$(C*k)(p) = \sum_{i+j=p} C(i) \cdot k(j) \tag{12}$$

After defined this convolution operator $*$, let $d$ be a dilation factor and let $*_d$ be equivalent to

$$(C *_d k)(p) = \sum_{i+d \cdot j=p} C(i) \cdot k(j) \tag{13}$$

where $*_d$ is served as the dilated convolution or a $d$- dilated convolution. Particularly, the common discrete convolution $*$ can be regarded as the $1$- dilated convolution. Setting the size of convolutional kernel with 3×3 as an example, and let $k_l$ be the discrete 3×3 convolution filters. Consider applying the filters with exponentially increasing dilation as

$$R_{l+1} = R_l *_\phi k_l \tag{14}$$

where $l = 0, 1, \ldots, n-2$, $\phi = 2^l$ and $R_l$ represents the size of the receptive field. The common convolution receptive field has a linear correlation with the layer depth, in that the receptive field size: $R_l^c = (2l+1) \times (2l+1)$. By contrast, the dilated convolution receptive field has an exponential correlation with the layer depth, where the receptive field size: $R_l^d = (2^{l+1}-1) \times (2^{l+1}-1)$. For instance, when $l = 4$, $R_l^c = 9 \times 9$ while $R_l^d = 31 \times 31$ with the same layer depth. Figure 2 illustrates the dilated convolution receptive field size, where: (a) corresponds to the 1-dilated convolution, which is equivalent to the common convolution operation at this point; (b) corresponds to the 2-dilated convolution; and (c) corresponds to the 4-dilated convolution.

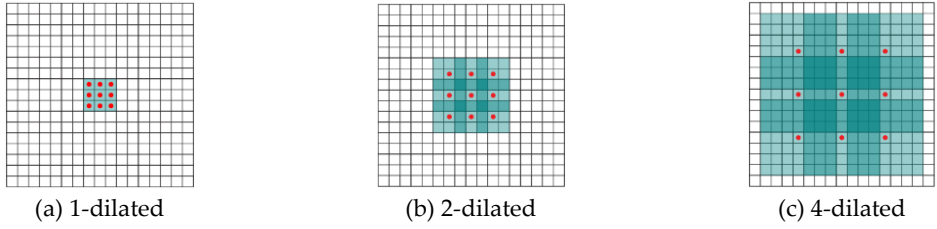

(a) 1-dilated      (b) 2-dilated      (c) 4-dilated

**Figure 2. receptive field size of dilated convolution. ($d$ =1, 2 and 4)**

In the proposed SAR-DRN model, considering that trade-off between feature extraction ability and reducing training time, the dilation factors of the 3×3 dilated convolutions from layer 1 to layer 7 are respectively set to 1, 2, 3, 4, 3, 2, and 1, empirically. Compared with other deep neural networks, we propose a lightweight model with only seven dilated convolution layers, as shown in Figure 3.

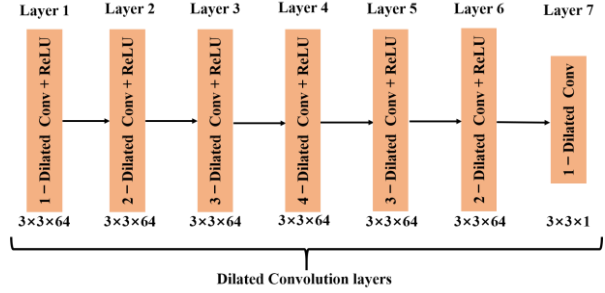

**Figure 3. Dilated convolution in the proposed model.**

*3.2 Skip Connections*

Although the increase of network layer depth can help to obtain more data feature expressions, it often results in the vanishing gradient problem, which makes the training of the model much harder. To solve this problem, a new structure called skip connection [34] has been created for the DCNNs, to obtain better training results. The skip connection can pass the previous layer's feature information to its posterior layer, maintaining the image details and avoiding or reducing the vanishing gradient problem. For the $l$-th layer, let $L^{(l)}$ be the input data, and let $f(L^{(l)},\{W,b\})$ be its feed-forward propagation with trainable parameters. The output of $(l+k)$-th layer with $k$-interval skip connection is recursively defined as follows:

$$L^{(l+k)} = f(L^{(l)},\{W,b\}_{l+1 \to l+k}) + L^{(l)} \tag{15}$$

For clarity, in the proposed SAR-DRN model, two skip connections are employed to connect layer 1 to layer 3 (as shown in Figure 4(a)) and layer 4 to layer 7(as shown in Figure 4(b)), whose effects are compared with no skip connections in discussion section.

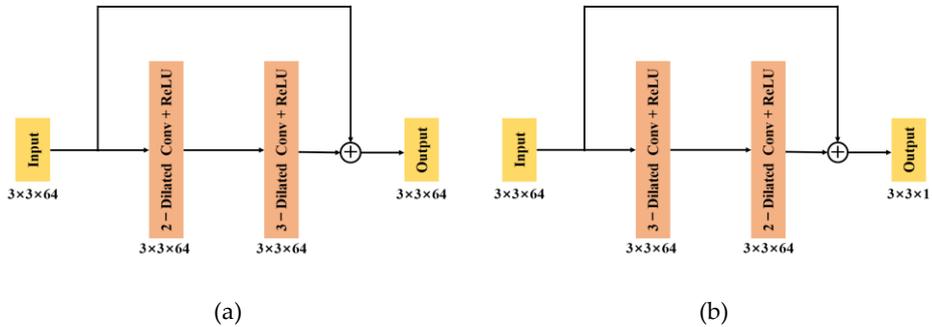

(a)      (b)

**Figure. 4. Diagram of skip connection structure in the proposed model. (a) connecting dilated convolution layer 1 to dilated convolution layer 3. (b) dilated convolution layer 4 to dilated convolution layer 7.**

*3.3 Residual Learning*

Compared with traditional data mapping, He *et al*. [35] found that residual mapping can acquire a more effective learning effect and rapidly reduce the training loss after passing through a multi-layer network, which has achieved state-of-the-art performance in object detection [36], image super-resolution [37] and so on. Essentially, Szegedy *et al.* [38] demonstrated that residual networks take full advantage of identity shortcut connections, which can efficiently transfer various levels of feature information between not directly connected layers without attenuation. In the proposed SAR-DRN model, the residual image $\varphi$ is defined as follows:

$$\varphi = y_i - x_i \tag{16}$$

As layer depth increasing, the degradation phenomenon manifests that common deep networks might have difficulties in approximating identical mappings by stacked non-linear layers like *Conv-BN-ReLU* block. By contrast, it is reasonable to consider that most pixel values in residual image $\varphi$ are very close to zero, and the spatial distribution of the residual feature maps should be very sparse, which can transfer the gradient descent process to a much smoother hyper-surface of loss to filtering parameters. Thus, searching for an allocation which is on the verge of the optimal for the network's parameters becomes much quicker and easier, allowing us to add more trainable layers to the network and improve its performance. The learning procedure with residual unit is easier to approximate to the original multiplicative speckle noise through the deeper and intrinsic non-linear feature extraction and expression, which can better weaken the range difference between optical images and SAR images.

Specifically for the proposed SAR-DRN, we choose a collection of $N$ training image pairs $\{x_i, y_i\}_N$ from the training data sets as described in 4.1 below, where $y_i$ is the speckled image, and $\theta$ is the network parameters. Our model uses the mean squared error (MSE) as the loss function:

$$loss(\Theta) = \frac{1}{2N}\sum_{i=1}^{N}\|\phi(y_i,\theta) - \varphi\|_2^2 \tag{17}$$

In summary, with the dilated convolution, skip connections and residual learning structure, the flowchart of learning a deep network for the SAR image despeckling process is described in Figure 5. To learn the complicated non-linear relation between the speckled image $y$ and original image $x$, the proposed SAR-DRN model is employed with converged loss between the residual image $\varphi$ and the output $\phi(y,\theta)$, then preparing for real speckle SAR image processing as illuminated in Figure 5.

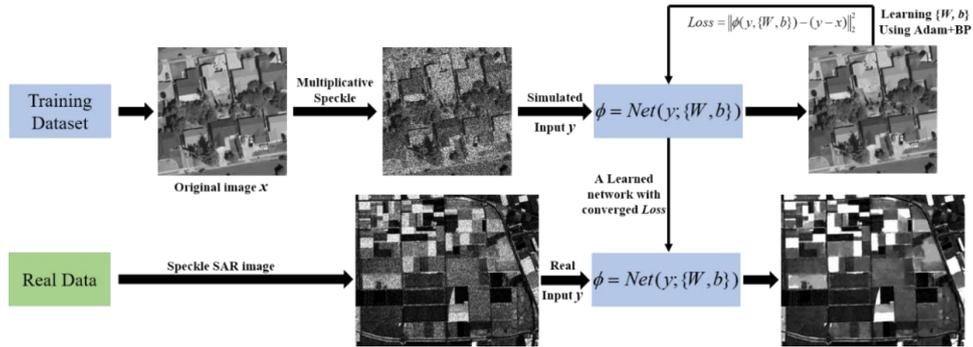

**Figure 5. The framework of SAR image despeckling based on deep learning.**

## 4. Experimental Results and Analysis

*4.1 Implementation Details*

**1) Training and Test Datasets**

Considering that it's quite hard to obtain clean reference training SAR images without speckle at all, we used the *UC Merced* land-use dataset [39] as our training dataset with different numbers of looks for simulating SAR image despeckling, which contains 21 scene classes with 100 images per class. By the reason that optical images and SAR images are statistically different, the amplitude information of optical images are processed before training for single-polarization SAR data despeckling, to better accord with the data distribution property of SAR images. To train the proposed SAR-DRN, we chose 400 images of size 256×256 from this dataset and set each patch size as 40×40 and stride equal to 10. Then 193,664 patches are cropped for training SAR-DRN with batch size as 128 for parallel computing. And the number of looks $L$ was set to noise levels of 1, 2, 4, and 8 for adding multiplicative speckle noise, respectively.

To test the performance of the proposed model, three examples of the Airplanes, Buildings and Rivers classes were respectively set up as simulated images. And for the real SAR image despeckling experiments, we used the classic *Flevoland* SAR image (cropped to 500×600), *Deathvalley* SAR image (cropped to 600×600), and *San Francisco* SAR image (cropped to 400×400) which are commonly used in real SAR data image despeckling.

**2) Parameter Setting and Network Training**

Table 1 lists the network parameters of each layer for SAR-DRN. The proposed model was trained using the Adam [40] algorithm as the gradient descent optimization method, with momentum $\beta_1 = 0.9$, momentum $\beta_2 = 0.999$, and $\varepsilon = 10^{-8}$, where the learning rate $\alpha$ was initialized to 0.01 for the whole network. The optimization procedure is given as below.

$$m_t = \beta_1 \cdot m_{t-1} + (1-\beta_1) \cdot \frac{\partial L}{\partial \theta_t} \tag{18}$$

$$n_t = \beta_2 \cdot n_{t-1} + (1-\beta_2) \cdot (\frac{\partial L}{\partial \theta_t})^2 \tag{19}$$

$$\Delta \theta_t = -\alpha \cdot \frac{m_t}{\sqrt{n_t} + \varepsilon} \tag{20}$$

where $\theta$ is the trainable parameter in the network of *t*-th iteration. The training process of SAR-DRN took 50 epochs (about 1,500 iterations), and after every 10 epochs, the learning rate was reduced through being multiplied by a descending factor *gamma* = 0.5. We used *Caffe* [41] framework to train the proposed SAR-DRN in the Windows 7 environment, 16 GB-RAM, with an Nvidia Titan-X (Pascal) GPU. The total training time costs about 4 hours 30 minutes, which is less than SAR-CNN [28] with about 9h 45 minutes under the same computational environment.

Table 1. The network configuration of the SAR-DRN model

|  | Configurations |
| --- | --- |
| **Layer 1** | Dilated Conv + ReLU: $64 \times 3 \times 3$, dilate=1, stride=1, pad=1 |
| **Layer 2** | Dilated Conv + ReLU: $64 \times 3 \times 3$, dilate=2, stride=1, pad=2 |
| **Layer 3** | Dilated Conv + ReLU: $64 \times 3 \times 3$, dilate=3, stride=1, pad=3 |
| **Layer 4** | Dilated Conv + ReLU: $64 \times 3 \times 3$, dilate=4, stride=1, pad=4 |
| **Layer 5** | Dilated Conv + ReLU: $64 \times 3 \times 3$, dilate=3, stride=1, pad=3 |
| **Layer 6** | Dilated Conv + ReLU: $64 \times 3 \times 3$, dilate=2, stride=1, pad=2 |
| **Layer 7** | Dilated Conv: $1 \times 3 \times 3$, dilate=1, stride=1, pad=1 |

**3) Compared Algorithms and Quantitative Evaluations**

To verify the proposed method, we compared SAR-DRN method with four mainstream despeckling methods: The probabilistic patch-based (PPB) filter [13] based on patch matching, SAR-BM3D [14] based on 3-D patch matching and wavelet, SAR-POTDF [16] based on sparse representation, and SAR-CNN [28] based on deep neural network. In the simulated-image experiments, the peak signal-to-noise ratio (PSNR) and structural similarity (SSIM) were employed as the quantitative evaluation indexes. And in the real-image experiments, the *ENL* was considered as the smoothness of a homogeneous region after SAR image despeckling (the *ENL* is commonly regarded as the quantitative evaluation index for real SAR image despeckling experiments), whose value is larger demonstrating the homogeneous region is smoother, as defined in Equation (3).

*4.2 Simulated-Data Experiments*

To verify the effectiveness of the proposed SAR-DRN model in SAR image despeckling, four different speckle noise levels of looks $L$=1, 2, 4 and 8 were set up for the three simulated images for PPB, SAR-BM3D, SAR-POTDF, SAR-CNN and ours. The PSNR and SSIM evaluation indexes and their standard deviations of the 10 times simulated experiments with the three images are listed in Tables 2 ,3 and 4, respectively, where the best performance is marked in bold.

**Table 2. Mean and Stand Deviation Results of PSNR (dB) and SSIM for Airplane with $L$=1, 2, 4, and 8**

| Looks | index | PPB | SAR-BM3D | SAR-POTDF | SAR-CNN | SAR-DRN |
|---|---|---|---|---|---|---|
| $L$=1 | PSNR | 20.11 ± 0.065 | 21.83 ± 0.051 | 21.75 ± 0.061 | 22.06 ± 0.053 | **22.97 ± 0.052** |
| | SSIM | 0.512 ± 0.001 | 0.623 ± 0.003 | 0.604 ± 0.003 | 0.623 ± 0.002 | **0.656 ± 0.001** |
| $L$=2 | PSNR | 21.72 ± 0.055 | 23.59 ± 0.062 | 23.79 ± 0.041 | 24.13 ± 0.048 | **24.54 ± 0.043** |
| | SSIM | 0.601 ± 0.001 | 0.693 ± 0.004 | 0.686 ± 0.003 | 0.710 ± 0.002 | **0.726 ± 0.002** |
| $L$=4 | PSNR | 23.48 ± 0.073 | 25.51 ± 0.079 | 25.84 ± 0.047 | 25.97 ± 0.051 | **26.52 ± 0.046** |
| | SSIM | 0.678 ± 0.003 | 0.755 ± 0.002 | 0.752 ± 0.002 | 0.748 ± 0.003 | **0.763 ± 0.002** |
| $L$=8 | PSNR | 24.98 ± 0.084 | 27.17 ± 0.064 | 27.56 ± 0.060 | 27.89 ± 0.062 | **28.01 ± 0.058** |
| | SSIM | 0.743 ± 0.003 | 0.800 ± 0.003 | 0.794 ± 0.004 | 0.801 ± 0.002 | **0.819 ± 0.003** |

**Table 3. Mean and Stand Deviation Results of PSNR (dB) and SSIM for Building with $L$=1, 2, 4, and 8**

| Looks | index | PPB | SAR-BM3D | SAR-POTDF | SAR-CNN | SAR-DRN |
|---|---|---|---|---|---|---|
| $L$=1 | PSNR | 25.05 ± 0.036 | 26.14 ± 0.059 | 25.10 ± 0.035 | 26.25 ± 0.052 | **26.80 ± 0.044** |
| | SSIM | 0.715 ± 0.002 | 0.786 ± 0.005 | 0.731 ± 0.001 | 0.775 ± 0.002 | **0.796 ± 0.003** |
| $L$=2 | PSNR | 26.36 ± 0.064 | 27.95 ± 0.046 | 27.44 ± 0.041 | 27.98 ± 0.058 | **28.39 ± 0.045** |
| | SSIM | 0.778 ± 0.003 | 0.831 ± 0.004 | 0.811 ± 0.003 | 0.826 ± 0.003 | **0.838 ± 0.002** |
| $L$=4 | PSNR | 28.05 ± 0.053 | 29.84 ± 0.033 | 29.56 ± 0.066 | 29.96 ± 0.057 | **30.14 ± 0.048** |
| | SSIM | 0.833 ± 0.002 | **0.879 ± 0.002** | 0.866 ± 0.002 | 0.869 ± 0.003 | 0.870 ± 0.002 |
| $L$=8 | PSNR | 29.50 ± 0.069 | 31.36 ± 0.070 | 31.55 ± 0.051 | 31.63 ± 0.054 | **31.78 ± 0.058** |
| | SSIM | 0.871 ± 0.00 | **0.902 ± 0.001** | 0.900 ± 0.002 | 0.901 ± 0.002 | 0.901 ± 0.001 |

**Table 4. Mean and Stand Deviation Results of PSNR (dB) and SSIM for Highway with $L$=1, 2, 4, and 8**

| Looks | index | PPB | SAR-BM3D | SAR-POTDF | SAR-CNN | SAR-DRN |
|---|---|---|---|---|---|---|
| $L$=1 | PSNR | 20.13 ± 0.059 | 21.12 ± 0.031 | 20.63 ± 0.047 | 21.07 ± 0.036 | **21.71 ± 0.024** |
| | SSIM | 0.472 ± 0.002 | 0.558 ± 0.002 | 0.530 ± 0.002 | 0.552 ± 0.003 | **0.613 ± 0.003** |
| $L$=2 | PSNR | 21.40 ± 0.073 | 22.62 ± 0.028 | 22.51 ± 0.063 | 22.88 ± 0.062 | **22.96 ± 0.057** |
| | SSIM | 0.572 ± 0.002 | **0.646 ± 0.002** | 0.637 ± 0.003 | 0.641 ± 0.002 | 0.644 ± 0.003 |
| $L$=4 | PSNR | 22.61 ± 0.037 | 24.29 ± 0.049 | 24.39 ± 0.071 | 24.46 ± 0.061 | **24.64 ± 0.063** |
| | SSIM | 0.674 ± 0.002 | 0.765 ± 0.003 | 0.768 ± 0.004 | 0.762 ± 0.003 | **0.772 ± 0.002** |
| $L$=8 | PSNR | 24.90 ± 0.045 | 26.41 ± 0.075 | 26.37 ± 0.044 | 26.48 ± 0.058 | **26.53 ± 0.046** |
| | SSIM | 0.764 ± 0.005 | 0.834 ± 0.002 | 0.837 ± 0.002 | 0.834 ± 0.003 | **0.836 ± 0.002** |

As shown in Table 2, 3 and 4, the proposed SAR-DRN model obtains all the best PSNR results and nine of the twelve best SSIM results in the four noise levels. When *L*=1, the proposed method outperforms SAR-BM3D by about 1.1 dB/0.6 dB/0.6 dB for Airplane, Building and Highway image, respectively. When *L*=2 and 4, SAR-DRN outperforms PPB, SAR-POTDF, SAR-BM3D, and SAR-CNN by at least 0.5 dB/0.7 dB/0.3 dB and 0.4 dB/0.3 dB/0.2 dB for Airplane/Building/Highway, respectively. Compared with the traditional despeckling methods above, the proposed method shows superior performance over the state-of-the-art methods on both quantitative and visual assessments, especially for strong speckle noise.

Figure 6, Figure 7 and Figure 8 correspondingly show the filtered images for the Airplane/Building/Highway images contaminated by 2-look speckle, 4-look speckle and 4-look speckle, respectively. It can be clearly seen that PPB has a good speckle-reduction ability, but PPB simultaneously creates many texture distortions, especially around the edges of the airplane, building and highway. SAR-BM3D and SAR-POTDF perform better than PPB on both the Airplane, Building and Highway images, especially for strong speckle noise such as *L*=1, 2 or 4, which reveals an excellent speckle-reduction ability and local detail preservation ability. Furthermore, they generate fewer texture distortions, as shown in Figure 6, 7 and 8. However, SAR-BM3D and SAR-POTDF also simultaneously result in over-smoothing, to some degree, as they mainly concentrate on some complex geometric features. SAR-CNN also shows a good speckle-reduction ability and local detail preservation ability, but it introduces some radiation distortions in homogeneous regions. Compared with the other algorithms above, SAR-DRN achieves the best performance in speckle reduction, concurrently avoiding introducing radiation and geometric distortion. In addition, from the red boxes of the Airplane and Building images in Figure 6, 7 and 8, respectively, it can be clearly seen that SAR-DRN also shows the best local detail preservation ability, while the other methods either miss partial texture details or produce blurry results, to some extent.

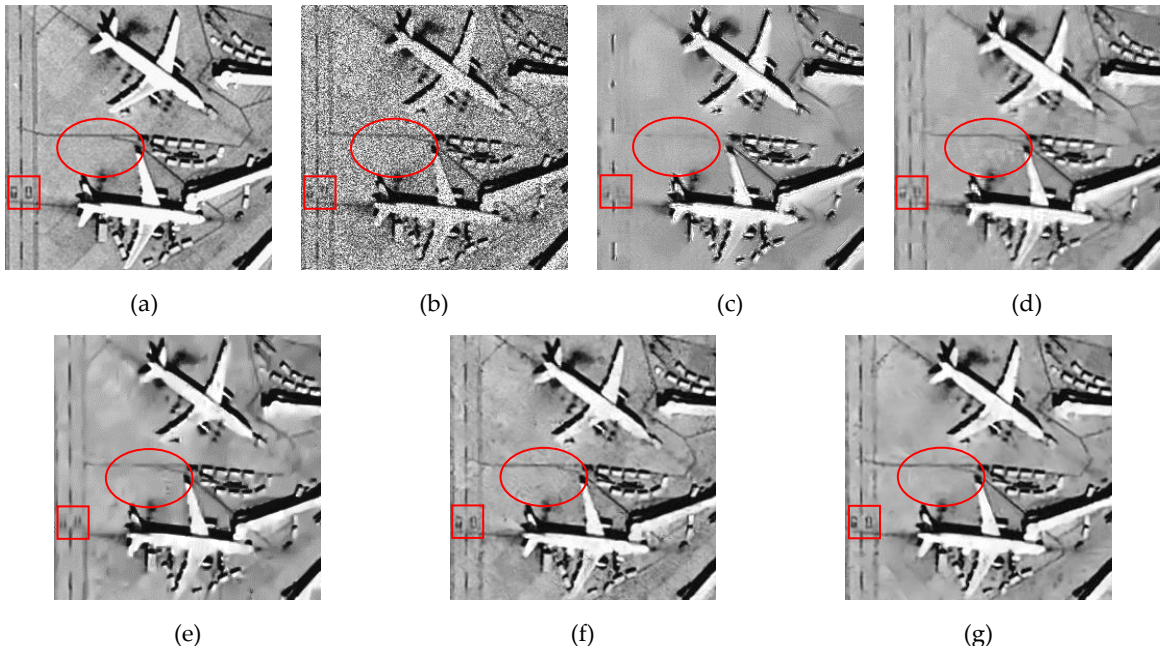

**Figure 6. Filtered images for the Airplane image contaminated by 2-look speckle. (a) Original image. (b) Speckled image. (c) PPB [13]. (d) SAR-BM3D [14]. (e) SAR-POTDF [16]. (f) SAR-CNN [28]. (g) SAR-DRN.**

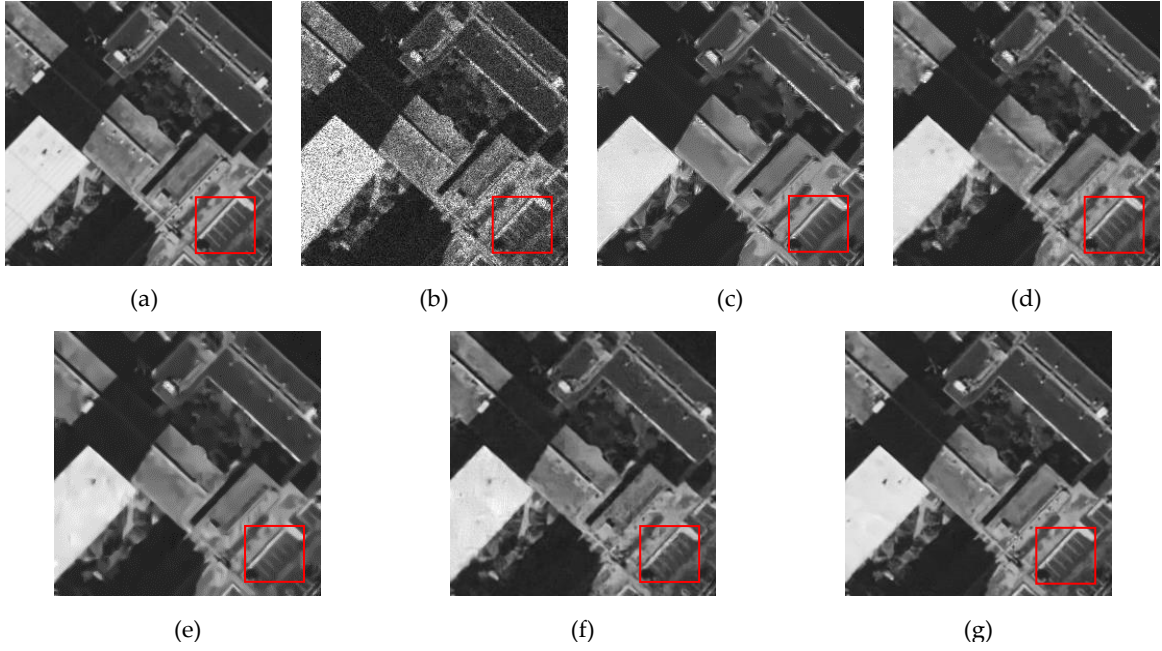

**Figure 7.** Filtered images for the Building image contaminated by 4-look speckle. (a) Original image. (b) Speckled image. (c) PPB [13]. (d) SAR-BM3D [14]. (e) SAR-POTDF [16]. (f) SAR-CNN [28]. (g) SAR-DRN.

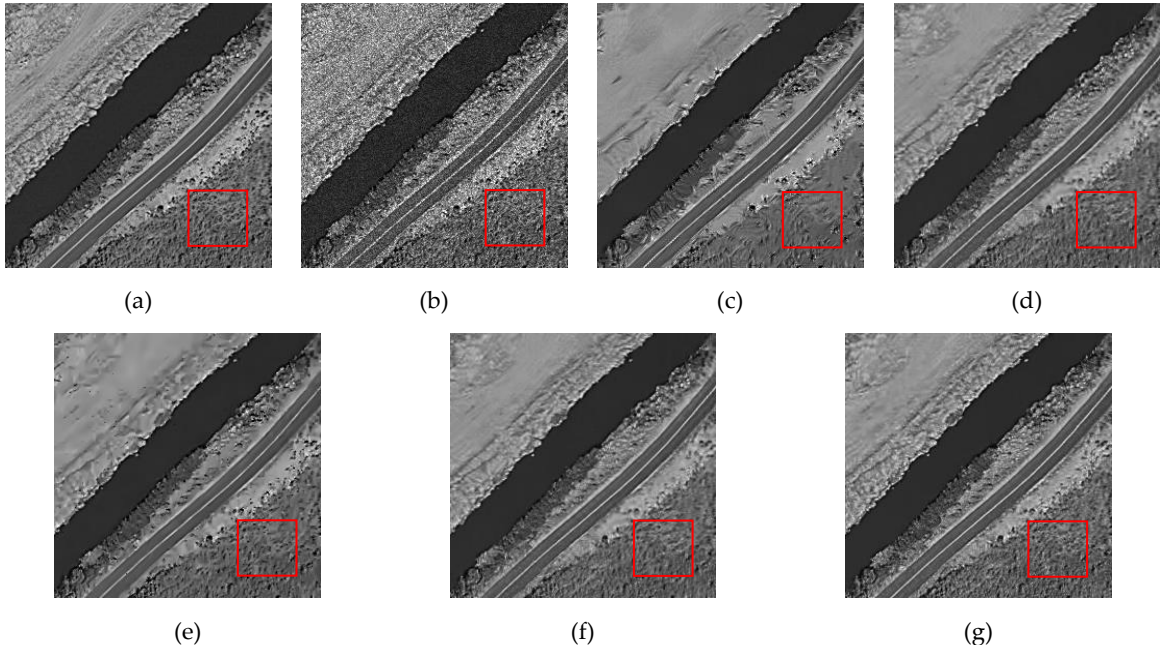

**Figure 8.** Filtered images for the Highway image contaminated by 4-look speckle. (a) Original image. (b) Speckled image. (c) PPB [13]. (d) SAR-BM3D [14]. (e) SAR-POTDF [16]. (f) SAR-CNN [28]. (g) SAR-DRN.

*4.3 Real-Data Experiments*

As shown in Figure 9, Figure 10 and Figure 11, we also compared the proposed method with the four state-of-the-art methods described above on three real SAR images. These three SAR images are all acquired by the Airborne Synthetic Aperture Radar(AIRSAR), which are all 4-look data. In Figure 9, it can be clearly seen that the result of SAR-BM3D still contains a great deal of residual speckle noise, while the results of PPB, SAR-POTDF, SAR-CNN, and the proposed SAR-DRN method reveal good speckle-reduction ability. PPB performs very well in speckle reduction, but it generates a few texture distortions in the edges of prominent objects. In homogeneous regions, SAR-POTDF does not perform as well in speckle reduction as the proposed SAR-DRN. As for SAR-CNN, its edge-preserving ability is weaker than that of SAR-DRN. Visually, SAR-DRN achieves the best performance in speckle reduction and local detail preservation, performing better than the other mainstream methods; In Figure 10, all the five methods can well reduce the speckle noise, but PPB obviously exists over-smoothing phenomenon; Besides, in Figure 11, the result of SAR-CNN still contains some residual speckle noise. Simultaneously, PPB, SAR-BM3D and SAR-POTDF also result in over-smoothing phenomenon, to some degree, as shown in the marked regions with complex geometric features. And it can be clearly seen that the proposed method has both well speckled noise reduction ability and preserving detail ability for the edge and texture information.

In addition, we also evaluated the filtered results, through *ENL* in Table 5 and EPD-ROA [15] in Table 6 to measure the speckle-reduction and edge-preserving ability, respectively. Because it's difficult to find homogeneous regions in Figure 11, the *ENL* values were respectively estimated from four chosen homogeneous regions of Figure 9 and Figure 10 (the red boxes in Figure 9(a) and Figure 10(a)). Clearly, SAR-DRN has a much better speckle-reduction ability than the other methods, which is consistent with the visual observation.

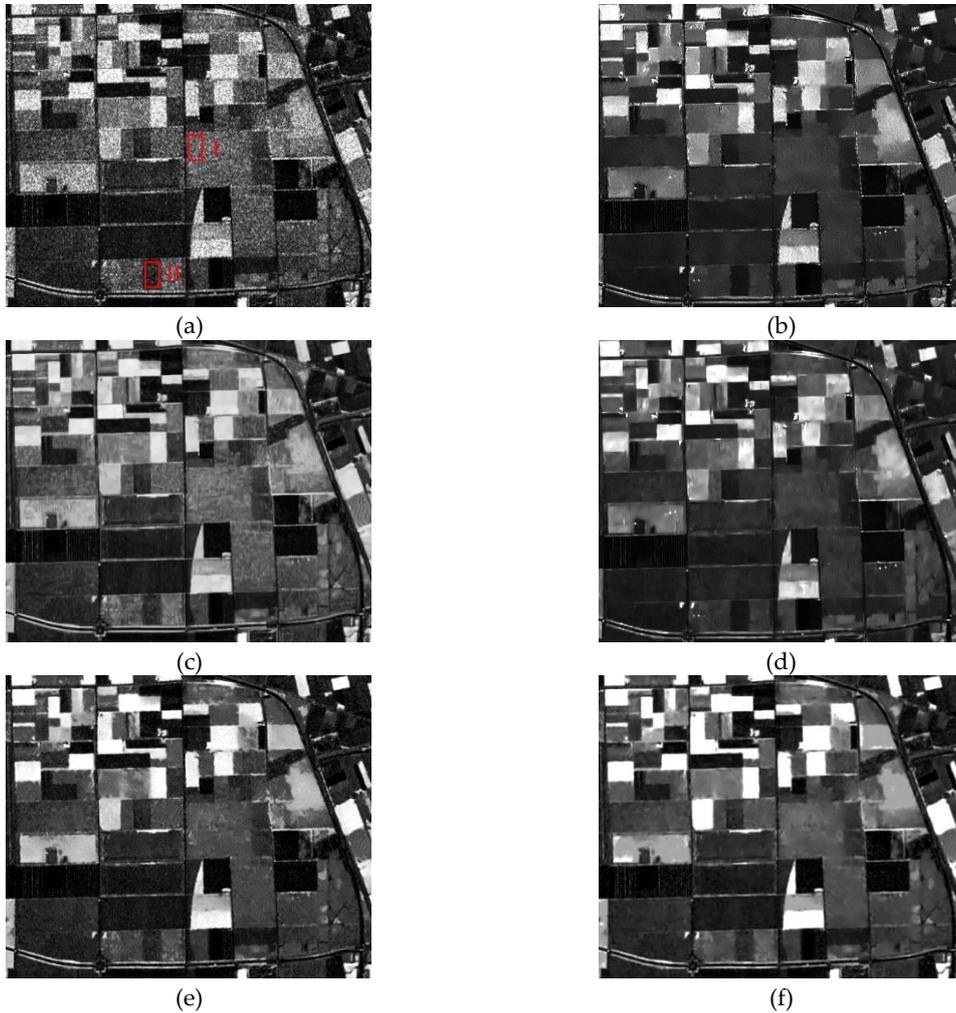

**Figure 9. Filtered images for the *Flevoland* SAR image contaminated by 4-look speckle. (a) Original image. (b) PPB [13]. (c) SAR-BM3D [14]. (d) SAR-POTDF [16]. (e) SAR-CNN [28]. (f) SAR-DRN.**

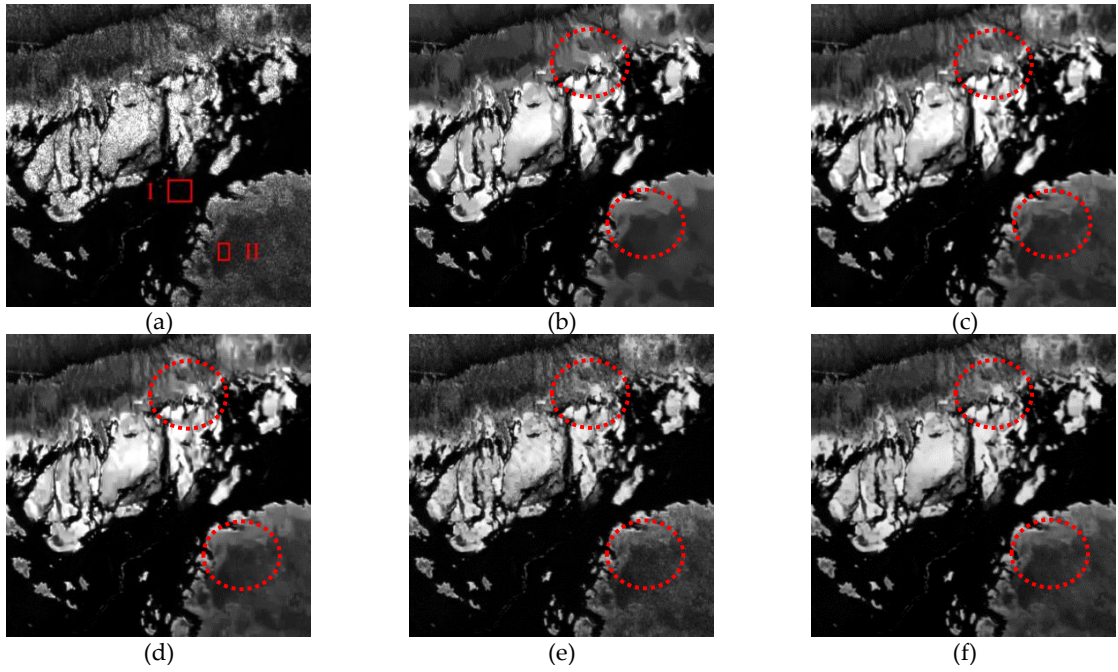

**Figure 10.** Filtered images for the *Deathvalley* SAR image contaminated by 4-look speckle. (a) Original image. (b) PPB [13]. (c) SAR-BM3D [14]. (d) SAR-POTDF [16]. (e) SAR-CNN [28]. (f) SAR-DRN.

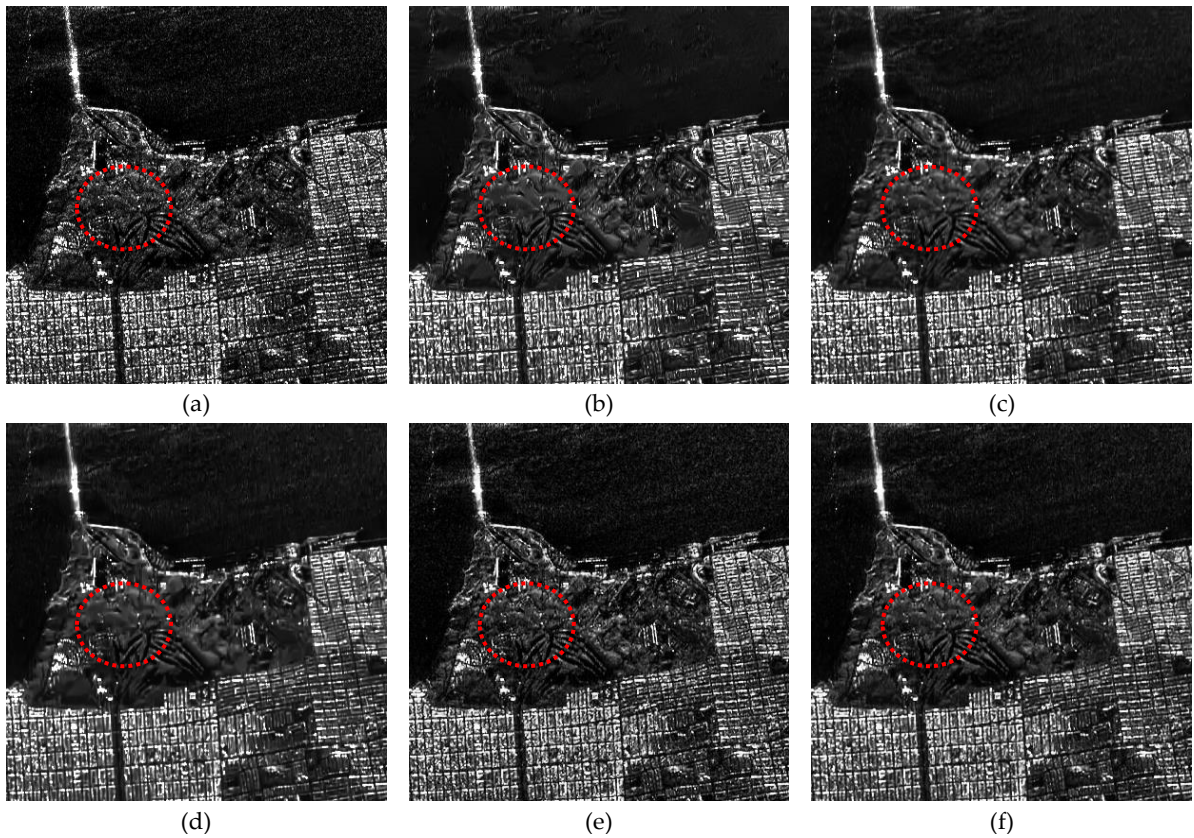

**Figure 11.** Filtered images for the *San Francisco* SAR image contaminated by 4-look speckle. (a) Original image. (b) PPB [13]. (c) SAR-BM3D [14]. (d) SAR-POTDF [16]. (e) SAR-CNN [28]. (f) SAR-DRN.

**Table 5. *ENL* results for the *Flevoland* and *Deathvalley* images**

| Data | | Original | PPB | SAR-BM3D | SAR-POTDF | SAR-CNN | SAR-DRN |
|---|---|---|---|---|---|---|---|
| **Figure 9** | Region I | 4.36 | 122.24 | 67.43 | 120.32 | 86.29 | **137.63** |
| | Region II | 4.11 | **56.89** | 24.96 | 38.90 | 23.38 | 45.64 |
| **Figure 10** | Region I | 5.76 | 14.37 | 12.65 | 12.72 | 13.26 | **14.58** |
| | Region II | 4.52 | 43.97 | **55.76** | 44.87 | 37.45 | 48.32 |

**Table 6. EPD-ROA indexes for the real despeckling results**

| Data | PPB | SAR-BM3D | SAR-POTDF | SAR-CNN | SAR-DRN |
|---|---|---|---|---|---|
| **Figure 9** | 0.619 | 0.733 | 0.714 | 0.748 | **0.754** |
| **Figure 10** | 0.587 | 0.714 | 0.702 | 0.698 | **0.723** |
| **Figure 11** | 0.632 | **0.685** | 0.654 | 0.621 | 0.673 |

*4.4 Discussion*

1) Dilated Convolutions and Skip Connections

As mentioned in Section III, dilated convolutions are employed in the proposed method, which can both enlarge the receptive field and maintain the filter size and layer depth with a lightweight structure. In addition, skip connections are also added to the despeckling model to maintain the image details and reduce the vanishing gradient problem. To verify the effectiveness of the dilated convolutions and skip connections, we implemented four sets of experiments in the same environment as that shown in Figure 12: 1) with dilated convolutions and skip connections (the red line); 2) with dilated convolutions but without skip connections (the green line); 3) without dilated convolutions but with skip connections (the blue line); and 4) without dilated convolutions and skip connections (the black line).

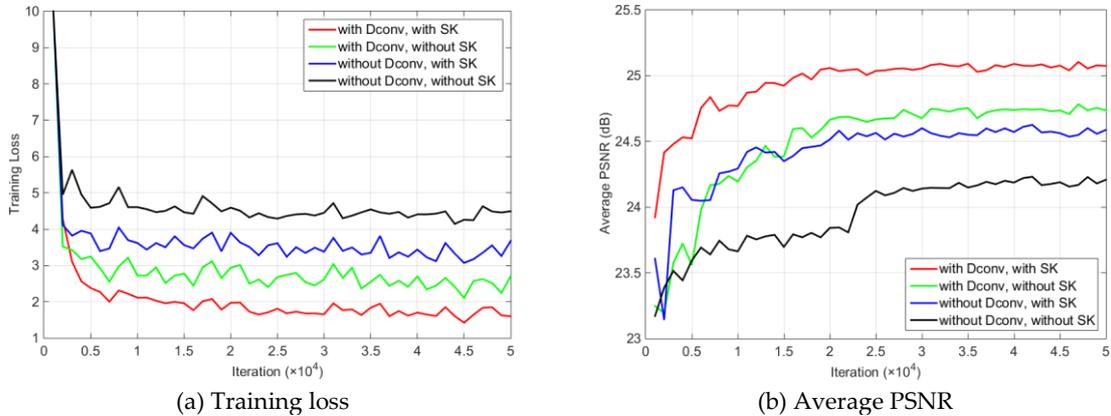

(a) Training loss    (b) Average PSNR

**Figure 12.** The simulated SAR image despeckling results of the four specific models in (a) training loss and (b) average PSNR, with respect to iterations. The four specific models were different combinations of dilated convolutions (Dconv) and skip connections (SK), and were trained with 1-look images in the same environment. The results were evaluated on the *Set14* [43] dataset.

As Figure 12 implies, the dilated convolutions can effectively reduce the training loss and enhance the despeckling performance (the less training Loss and the best PSNR), which also testifies that augmenting the contextual information through enlarging the receptive field is effective for recovery the degraded image, as demonstrated in Section III for dilated convolution. Meanwhile, the skip connections also accelerate the convergence speed of the network and enhance the model stability, as comparison with or without skip connection in Figure 12. Besides, the combination of dilated convolution and skip connections can promote each other's effect, up from about 1.1 dB in PSNR compared with the combination of without dilated convolution and without skip connections.

2) With or Without Batch Normalization (BN) in the Network

Unlike the methods proposed in [28] and [29], which utilize batch normalization to normalize the output features, SAR-DRN does not add this preprocessing layer, considering that the skip connections can also maintain the outputs of the data distribution in the different dilated convolution layers. The quantitative comparison of the two structures for SAR image despeckling is provided in Section IV. Furthermore, getting rid of the BN layers can simultaneously reduce the amount of computation, saving about 3 hours of training time in the same environment. Figure 13 shows that this modification improves the despeckling performance and reduces the complexity of the model. Regarding this phenomenon, we suggest that a probable reason is that the input and output have a highly similar spatial distribution for this regression problem, while the BN layers normalize the hidden layers' output, which destroys the representation of the original space [44].

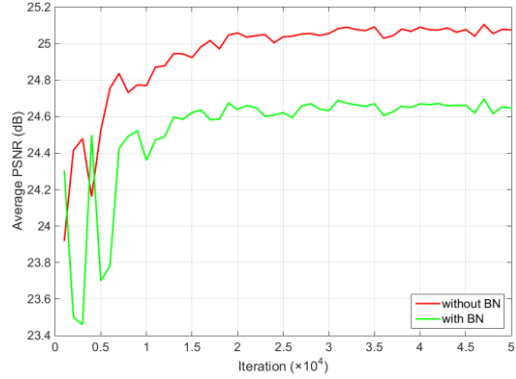

**Figure 13. The simulated SAR image despeckling results of the two specific models with/without batch normalization (BN). The two specific models were trained with 1-look images in the same environment, and the results were evaluated on the *Set14* [43] dataset.**

3) Runtime Comparisons

For evaluating the efficiency of despeckling algorithms, we make statistics of runtime under the same environment with MALAB R2014b, as listed in Table 7. Distinctly, SAR-DRN exhibits the lowest run-time complexity than other algorithms, because of the lightweight model with only 7 layers than other deep learning method like SAR-CNN [28] with 17 layers.

**Table 7. Runtime comparisons for five despeckling methods with an image of size 256 × 256 (Seconds)**

| Method  | PPB   | SAR-BM3D | SAR-POTDF | SAR-CNN | Ours     |
|---------|-------|----------|-----------|---------|----------|
| Runtime | 10.13 | 16.48    | 12.83     | 1.13    | **0.38** |

**5. Conclusion**

In this paper, we have proposed a novel deep learning approach for the SAR image despeckling task, learning an end-to-end mapping between the noisy and clean SAR images. Differently from common convolutions operation, the presented approach is based on dilated convolutions, which can both enlarge the receptive field and maintain the filter size with a lightweight structure. Furthermore, skip connections are added to the despeckling model to maintain the image details and avoid the vanishing gradient problem. Compared with the traditional despeckling methods, the proposed SAR-DRN approach shows a state-of-the-art performance in both simulated and real SAR image despeckling experiments, especially for strong speckle noise.

In our future work, we will investigate more powerful learning models to deal with the complex real scenes in SAR images. Considered that the training of our current method performed for each number of looks, we will explore an integrated model to solve this problem. Furthermore, the proposed approach will be extended to polarimetric SAR image despeckling, whose noise model is much more complicated than that of single-polarization SAR. Besides, for better reducing speckle noise in more complex real SAR image data, some *prior* constraint like multi-channel patch matching, band selection, location *prior* and locality adaptive discriminant analysis [45-48], can also be considered for improve precision of despeckling results. In addition, we will try to collect enough SAR images and then train the model with multi-temporal data [49] for SAR image despeckling, which will be sequentially explored to the future studies.

**Acknowledgments:** This work was supported by grants from the National Natural Science Foundation of China under Grants 61671334.

**Author Contributions:** Qiang Zhang proposed the method and performed the experiments; Qiang Zhang, Qiangqiang Yuan., Jie Li. and Zhen Yang conceived and designed the experiments; Qiang Zhang, Qiangqiang Yuan., Jie Li. Zhen Yang and Xiaoshuang Ma wrote the manuscript. All the authors read and approved the final manuscript.

**Conflicts of Interest:** The authors declare no conflict of interest.